\documentclass[11pt]{article}

\usepackage[final]{acl}
\usepackage{times}
\usepackage{latexsym}

\usepackage[T1]{fontenc}
\usepackage{xurl}
\usepackage[utf8]{inputenc}

\usepackage{microtype}

\usepackage{inconsolata}

\usepackage{graphicx}
\usepackage[inline]{enumitem}
\usepackage{tabularx}
\usepackage{booktabs}
\usepackage{listings}
\usepackage{placeins}

\usepackage{xcolor}
\title{Temporal Leakage in Search-Engine Date-Filtered Web Retrieval: A Retrospective Forecasting Case Study}
\author{
  \textbf{Ali El Lahib}\textsuperscript{1},
  \textbf{Ying-Jieh Xia}\textsuperscript{1},
  \textbf{Zehan Li}\textsuperscript{2},
  \textbf{Yuxuan Wang}\textsuperscript{1},
  \textbf{Xinyu Pi}\textsuperscript{1} \\
  \textsuperscript{1}University of California, San Diego \\
  \textsuperscript{2}University of Chicago \\
  \texttt{\{aellahib, yix050, waw009, xpi\}@ucsd.edu} \\
  \texttt{zehan@uchicago.edu}
}

\begin{document}
\maketitle

\begin{abstract}
Search-engine date filters are widely used to enforce pre-cutoff retrieval in retrospective evaluations of search-augmented forecasters. We show this approach is unreliable across two major search engines: auditing Google Search's \texttt{before:} filter and DuckDuckGo's date-range filter, we find that at least one retrieved page contains major post-cutoff leakage for 71\% of questions on Google and 81\% on DuckDuckGo, and the answer is directly revealed for 41\% and 55\%, respectively. Using \texttt{gpt-oss-120b} to forecast with these leaky documents, we demonstrate inflated prediction accuracy (Brier score 0.10 vs.\ 0.24 with leak-free documents). We characterize recurring leakage mechanisms, including updated articles, related-content modules, unreliable metadata, and absence-based signals, and argue that date-restricted search on these engines is insufficient for credible retrospective evaluation. We recommend stronger retrieval safeguards or evaluation on frozen, time-stamped web snapshots.
\end{abstract}

\section{Introduction}
\begin{figure}[t]
  \includegraphics[width=\columnwidth]{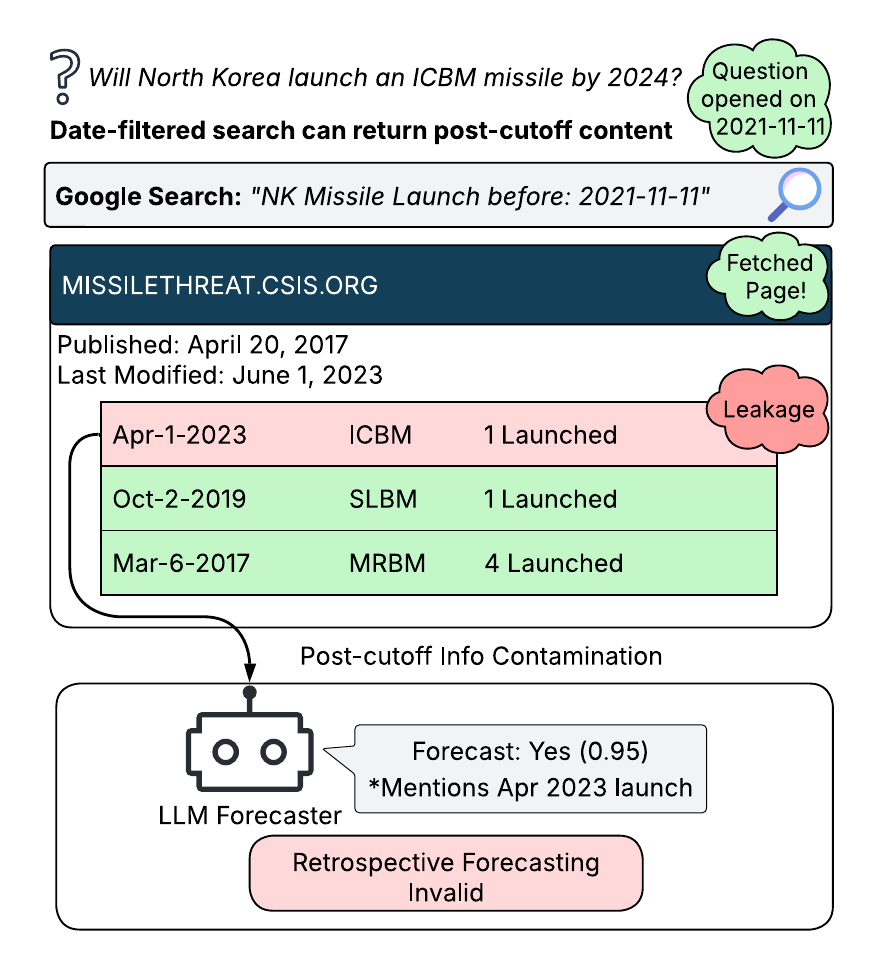}
  \caption{Date-filtered search retrieves a page updated after the question open date, leaking post-cutoff evidence into the LLM’s forecast. This inflates apparent performance and invalidates retrospective evaluation.}
  \label{fig:leakage_example}
  \vspace{-4mm}
\end{figure}

\textbf{Retrospective forecasting (RF)} evaluates forecasting systems on questions whose outcomes are already known. This setup requires that evidence available to the forecaster predates the resolution of each question. Without this guarantee, post-resolution information can leak into the retrieved documents, artificially inflating accuracy and undermining the validity of the evaluation (Figure~\ref{fig:leakage_example}).

Forecasting future events is a critical task for decision-making in policy, business, and science. Recent work has explored whether large language models (LLMs) can match or exceed human forecasters \citep{halawi2024approaching, phan2024llmssuperhuman, hsieh2024reasoning}, with some systems achieving near-human performance on \citet{metaculus2025platform} and other competitive forecasting platforms. Unlike most NLP tasks where static test sets suffice, evaluating forecasting ability poses a distinct challenge: ground-truth labels are only observed once future events resolve, which may take months or years. RF sidesteps this delay by backtesting on resolved questions while enforcing an information cutoff that restricts evidence to what was available at the time of prediction. This enables rapid iteration and immediate quantitative feedback.
In practice, most RF pipelines enforce the cutoff using search-engine date filters or by filtering on reported publication timestamps \citep{halawi2024approaching, phan2024llmssuperhuman, hsieh2024reasoning}. The same approach appears in related time-sensitive retrieval tasks, including claim verification \citep{AVeriTeC_Schlichtkrull}, dynamic fact-checking \citep{braun2025defamedynamicevidencebasedfactchecking}, and timeline summarization \citep{wu-etal-2025-unfolding}. The assumption is intuitive: filtering results by date should exclude documents published/updated after the cutoff, preventing post-cutoff facts from entering the retrieval.
Some prior work has noted that date-filtered search may not perfectly exclude post-cutoff content \citep{paleka2025pitfallsevaluatinglanguagemodel, futuresearch2025benchfuturepastcastingbenchmark}. However, since these claims rely on a few hand-picked examples, it remains unclear whether leakage is a rare edge case or a systematic problem, and how much leakage affects downstream forecasting accuracy. In this paper, we provide the first systematic study of search-engine date filtering for RF. We audit Google Search's \texttt{before:} filter across 393 resolved forecasting questions and DuckDuckGo's date-range filter across 389. Across 38,879 and 34,454 retrieved pages respectively, we find leakage is pervasive, not incidental. We further demonstrate that this leakage substantially inflates measured forecasting performance, and we characterize the mechanisms by which post-cutoff information enters date-filtered results.
\noindent We make three contributions:
\begin{itemize}[itemsep=0.5pt, topsep=1pt, leftmargin=*]
\item \textbf{Leakage Audit.} We audit Google's \texttt{before:} date filter and DuckDuckGo's date-range filter across \textasciitilde390 resolved forecasting questions and find leakage is pervasive. At least one retrieved page contains major post-cutoff information for 71\% of questions on Google and 81\% on DuckDuckGo, and at least one page directly reveals the answer for 41\% and 55\%, respectively.

\item \textbf{Downstream Impact.} We measure the effect on forecasting accuracy by comparing LLM predictions with and without leaked documents, demonstrating a large misleading performance gain (Brier score 0.10 vs. 0.24) \citep{brier1950verification}.
\item \textbf{Leakage Mechanisms.} We identify and categorize recurring pathways by which post-cutoff information enters date-filtered results, including updated article content, related-content modules, unreliable metadata, and absence-based signals.
\end{itemize}

Our findings demonstrate that date-restricted search is insufficient for credible retrospective evaluation. We recommend stronger retrieval safeguards or evaluation on frozen, time-stamped web snapshots. While our experiments focus on RF, the failure mode is general: pipelines that treat search-engine date filters as sufficient to prevent post-cutoff information from entering the retrieval might be vulnerable to the same leakage problem. We release the questions used, generated queries, retrieved URLs, and the audit time period at \url{https://github.com/theolivecode/WebDataLeakageAudit}.

\section{Related Work}
\paragraph{LLM Forecasting.} A growing body of work evaluates whether LLMs can serve as effective forecasters. \citet{halawi2024approaching, phan2024llmssuperhuman, hsieh2024reasoning} benchmark LLM predictions against human forecasters on forecasting platforms \citep{metaculus2025platform, manifold_markets}, using retrieval-augmented approaches that provide models with web-sourced evidence. These studies enforce information cutoffs via date-filtered search or publication timestamps, assuming this prevents post-cutoff information from entering retrieval. Fixed historical corpora such as Autocast \citep{zou2022forecastingfutureworldevents} offer strong temporal control but are limited for evaluating modern LLMs, whose knowledge cutoffs often cover the questions in these corpora.
\paragraph{Retrospective vs. Prospective Evaluation.} RF enables rapid, large-scale evaluation but introduces the risk of information leakage. Prospective benchmarks such as ForecastBench \citep{karger2025forecastbench} and FutureX \citep{zeng2025futurexadvancedlivebenchmark} address this by evaluating on unresolved questions in real time, though at the cost of slower iteration due to waiting for questions to resolve. FutureSearch \citep{futuresearch2025benchfuturepastcastingbenchmark} proposes an intermediate approach using frozen web snapshots collected before question resolution. Their system uses live Google search to rank results but filters them to return only pages stored in their pre-resolution snapshot database. While this constrains the content a forecaster can access, the authors acknowledge that live search ranking may still introduce bias, as the ordering of results reflects present-day relevance signals rather than those available at the cutoff date. Our findings provide empirical support for moving away from live date-filtered search, validating the motivation behind snapshot-based approaches.
\paragraph{Concerns About Date-Filtered Search.} \citet{paleka2025pitfallsevaluatinglanguagemodel} raise qualitative concerns about the reliability of search-engine date filters for retrospective evaluation, noting that web pages may be updated after publication, metadata may be missing or stale, and dynamic page components can introduce current information into otherwise historical content. Our work provides the systematic, quantitative analysis these observations call for. While some mechanisms we identify overlap with those noted by \citet{paleka2025pitfallsevaluatinglanguagemodel}, others, such as websites displaying incorrect self-reported timestamps and absence-based signals, are newly documented.

% \paragraph{Time-Sensitive Retrieval in Other Tasks.} Date-restricted retrieval is also used in dynamic fact-checking \citep{braun2025defamedynamicevidencebasedfactchecking} and timeline summarization \citep{wu-etal-2025-unfolding}. Our findings suggest these tasks face similar leakage risks when relying on search-engine date filters.

\section{Methodology}
We collected 393 resolved forecasting questions from tournaments hosted on the Metaculus platform \citep{metaculus2025platform}, spanning resolution dates from 2021 to 2025 (see Appendix~\ref{sec:example-question} for an example). For each question, we (i) generated 10 to 20 search queries using an LLM prompted with the question title and background, (ii) retrieved approximately 100 unique URLs per question via Google Search's \texttt{before:} filter and DuckDuckGo's date-range filter (start date 2000-01-01), with the cutoff set to the question's opening date on the forecasting platform, and (iii) fetched each page's content. DuckDuckGo returned usable results for 389 of the 393 questions. In total, we obtained 38,879 URLs on Google and 34,454 on DuckDuckGo for analysis (Table~\ref{tab:leakage_profile}). For pages exceeding 7,680 tokens, we applied Maximal Marginal Relevance \citep{10.1145/290941.291025} to select the most relevant passages. See Appendix~\ref{sec:prompt-query} for the query generation prompt and Appendix~\ref{sec:document-processing} for further processing details.

\begin{table}[t]
\centering
\small
\setlength{\tabcolsep}{4pt}
\begin{tabular*}{\columnwidth}{@{\extracolsep{\fill}}lrr@{}}
\toprule
\textbf{Metric} & \textbf{Google} & \textbf{DuckDuckGo} \\
\midrule
Questions evaluated       & 393             & 389             \\
URLs fetched              & 38,879          & 34,454          \\
URLs w/ post-cutoff info  & 33.2\% (12,903) & 34.5\% (11,898) \\
\midrule
\multicolumn{3}{@{}l}{\textit{Question-level leakage severity}} \\
Topical (score $\geq$ 1)      & 98.5\%        & 98.2\%          \\
Weak signal (score $\geq$ 2)  & 94.1\%        & 96.1\%          \\
Major signal (score $\geq$ 3) & 71.0\%        & 81.2\%          \\
Direct answer (score 4)       & 41.0\%        & 54.8\%          \\
\bottomrule
\end{tabular*}
\caption{Leakage prevalence on Google Search's \texttt{before:} filter and DuckDuckGo's date-range filter. Question-level severity rows report the percentage of questions with at least one retrieved URL at or above each leakage score.}
\label{tab:leakage_profile}
\vspace{-4mm}
\end{table}

\subsection{Leakage Severity Scoring}
\label{sec:scoring}
We developed a 0 to 4 severity scale to quantify post-cutoff information leakage, where post-cutoff information includes any event, data point, or entity that did not exist or was not public knowledge prior to the cutoff date. A score of 0 indicates no post-cutoff information or post-cutoff information irrelevant to the question; 1, topical but uninformative; 2, weak directional signal; 3, major signal enabling strong inference or decisive for a partial component; and 4, directly reveals the answer. Absence-based signals, where comprehensive sources omit expected information, are capped at 3 to avoid over-interpreting omissions. See Appendix~\ref{app:prompt-judge} for the complete scoring rubric and examples.

\subsection{LLM-as-a-Judge Implementation}
\label{sec:llm-judge}
To score leakage at scale, we implemented an LLM-as-a-Judge system \citep{zheng2023judging}. Each request includes the question title, background, resolution criteria, resolved answer, cutoff date (set to the question's opening date), webpage content, and the leakage scoring rubric with examples. The model outputs a JSON object containing: (i) whether the page contains post-cutoff information, (ii) a leakage score (0--4), and (iii) reasoning identifying specific post-cutoff content and justifying the score. We used \texttt{gpt-oss-120b} with temperature 0.5 \citep{openai2025gptoss120bgptoss20bmodel}. The prompt was developed iteratively, refined through inspection of model outputs at each severity level and the addition of in-context examples, with our human-annotated set used to validate prompt performance throughout (Section~\ref{sec:reliability}). See Appendix~\ref{app:prompt-judge} for the full LLM-as-a-Judge leakage scoring prompt.

\subsection{LLM Judge Reliability}
\label{sec:reliability}
Two annotators labeled disjoint subsets totaling 134 documents using the same rubric (with at least 19 examples per score level), then cross-reviewed each other's labels and resolved disagreements through discussion against the rubric until consensus was reached. LLM-human agreement reached 76.1\% exact accuracy (combining scores 0 and 1, as both indicate no actionable leakage) and 0.85 Quadratic Weighted Kappa (QWK), indicating disagreements typically occur between adjacent categories. The F1 score for direct leakage (score 4) was 0.82, confirming reliable detection of the most severe cases. See Appendix~\ref{sec:appendix-llm-human-agreement} for the full confusion matrix.

\subsection{Forecasting Experiment Setup}
\label{sec:forecast-setup}
% To measure downstream impact, we evaluated \texttt{gpt-oss-120b} on binary questions opened in 2025 (post-knowledge cutoff) with at least one score-4 document, comparing Brier scores when providing documents grouped by leakage level. This was run on Google-audited documents only.

% To measure downstream impact, we evaluated \texttt{gpt-oss-120b} on binary questions opened in 2025 with at least one score-4 document, comparing Brier scores~\citep{brier1950verification} when providing documents grouped by leakage level. Restricting to 2025 questions ensures baseline forecasting accuracy reflects genuine uncertainty rather than memorized answers, since these questions post-date the model's knowledge cutoff. We ran forecasting on the Google-audited documents only.

To measure downstream impact, we evaluated \texttt{gpt-oss-120b} on binary questions opened in 2025 with at least one score-4 document, comparing Brier scores when providing documents grouped by leakage level. We restrict to 2025 questions since they post-date the model's knowledge cutoff. This was run on Google-audited documents only.

\section{Results}

\subsection{Overall Leakage Prevalence}
\label{sec:prevalence}
Date filtering fails to prevent information leakage on either engine. As shown in Table~\ref{tab:leakage_profile}, 98.5\% of questions on Google and 98.2\% on DuckDuckGo return at least one URL containing topical post-cutoff information, demonstrating that neither the \texttt{before:} operator nor DuckDuckGo's date-range filter reliably filter content by actual information date. More critically, at least one document with major leakage (score $\geq$3) appears for 71.0\% of questions on Google and 81.2\% on DuckDuckGo, and at least one document directly reveals the answer (score 4) for 41.0\% and 54.8\%, respectively. In such cases, forecasting reduces to information retrieval rather than reasoning under uncertainty.

\subsection{Impact on Forecasting Performance}
\label{sec:performance}

% Critically, we examine whether detected leakage actually affects model predictions. We compare forecasting accuracy when models are provided with documents at different leakage levels.

\begin{table}[t]
\centering
\small
\setlength{\tabcolsep}{2pt}
% \begin{tabular}{lccc}
\begin{tabular*}{\columnwidth}{@{\extracolsep{\fill}}lccc@{}}

\toprule
\textbf{Retrieval Condition} & \textbf{Avg Sources} & \multicolumn{2}{c}{\textbf{Brier Score}}\\
\cmidrule(lr){3-4}
& & \textbf{Mean} & \textbf{Median}\\
\midrule
No retrieval (baseline) & -- & 0.244 & 0.090 \\
Score 0, no post-cutoff info & 73.5 & 0.242 & 0.102 \\
Scores 2–4 (weak to full) & 9.6 & 0.128 & 0.023\\
Scores 3–4 (strong to full) & 4.8 & \textbf{0.108} & \textbf{0.014}\\
Score 4 only (full leakage) & 2.6 & 0.129 & \textbf{0.014} \\
\bottomrule
\end{tabular*}
\caption{Forecasting performance by document leakage level on 93 binary questions from 2025 with at least one score-4 document. Lower Brier is better.}
\label{tab:brier}
\end{table}

We examine whether detected leakage affects model predictions by comparing forecasting accuracy across retrieval conditions on Google-audited documents (Table~\ref{tab:brier}). When the model received only leak-free documents (score 0, containing no post-cutoff info), performance matched the no-retrieval baseline with a Brier score of 0.24. For reference, predicting 50\% on every question yields 0.25. In contrast, providing strong and direct leakage documents (score $\geq$3, averaging 4.8 sources) reduced the Brier score to 0.108. Restricting to only score-4 documents (2.6 sources on average) yielded a slightly higher Brier score of 0.129. This difference reflects the value of corroboration. Additional score-3 documents provide context that helps the model interpret evidence more reliably and avoid overreacting to a single misleading or misread snippet. In both leaky settings, access to post-cutoff information substantially inflates apparent forecasting performance, undermining the validity of retrospective evaluations that rely on date-filtered search.

\subsection{Leakage Mechanisms}
\label{sec:mechanisms}

We identify four primary mechanisms through which post-cutoff information enters date-filtered results:

\paragraph{Direct Page Updates.} Pages are updated over time, introducing information that postdates the cutoff. For the question ``Will North Korea launch another intercontinental ballistic missile before 2024'' (open/cutoff date 2021-11-11), the date filter returned a missile tracking database (\texttt{missilethreat.csis.org}) first published in 2017, but the page had been continuously updated to include launch activities through 2023.

\paragraph{Related Content Leakage.} Sidebars and related articles sections can inject current content into otherwise historical pages. For the same ICBM question, the date filter returned a 2016 article whose main content contained no post-cutoff information. However, a related articles section on the page included a snippet about a December 2023 ICBM launch, fully revealing the answer despite the main article predating the cutoff.

\paragraph{Absence-Based Signal.} Sometimes the absence of expected information is meaningful, and might lead the LLM to conclude an answer. For the question ``Will there be a US-Iran war by 2024?'' with an open date on 2021-10-07, the date filter returned a CNN article containing a comprehensive US-Iran conflict timeline covering 1951--2025 with no mention of a war. This allows the model to reasonably infer the answer.

\paragraph{Unreliable Metadata.} Self-reported timestamps can be stale or incorrect, so post-filtering by scraped dates is not guaranteed to be reliable. For instance, for ``Will an additional state join NATO before 2024?'' (cutoff 2021-11-18), a retrieved page reports as last updated in 2020 yet contains text stating Finland joined NATO in 2023 and other facts in 2024, yielding a direct leakage (Score 4). This mechanism can bypass pipelines that double check retrieval by filtering on extracted publication or update dates. 

Full URLs for each case are listed in Appendix~\ref{sec:appendix-case-studies}.

\subsection{Temporal Variation in Leakage Severity}
\label{sec:temporal-variation-in-leakage-severity}

\begin{table}[t]
\centering
\small
\setlength{\tabcolsep}{3pt}
\begin{tabular*}{\columnwidth}{@{\extracolsep{\fill}}lrrrr@{}}
\toprule
\textbf{Cutoff} & \multicolumn{2}{c}{\textbf{Google}} & \multicolumn{2}{c}{\textbf{DuckDuckGo}} \\
\cmidrule(lr){2-3} \cmidrule(lr){4-5}
\textbf{Year}   & \textbf{Rate} & \textbf{URLs} & \textbf{Rate} & \textbf{URLs} \\
\midrule
2021            & 46.3\% & 1,831 / 3,955   & 47.1\% & 1,932 / 4,100  \\
2022            & 46.5\% & 3,115 / 6,703   & 48.0\% & 3,170 / 6,605  \\
2023            & 34.5\% & 2,008 / 5,821   & 31.4\% & 1,822 / 5,811  \\
2025            & 26.6\% & 5,949 / 22,400  & 27.7\% & 4,974 / 17,938 \\
\midrule
\textbf{Total}  & \textbf{33.2\%} & \textbf{12,903 / 38,879} & \textbf{34.5\%} & \textbf{11,898 / 34,454} \\
\bottomrule
\end{tabular*}
\caption{Percentage of URLs containing post-cutoff information by cutoff year. URLs column shows URLs with post-cutoff info over total URLs retrieved. The dataset did not contain questions opened in 2024.}
\label{tab:post_cutoff_by_year}
\end{table}
We further analyze the temporal dynamics of leakage severity by comparing results across cutoff years (2021--2023 and 2025). As shown in Table~\ref{tab:post_cutoff_by_year}, search results for the earliest questions (2021--2022) exhibit a consistently high density of post-cutoff information across both engines (>46\%). The 2023 cohort shows a notable drop (Google 34.5\%, DuckDuckGo 31.4\%), followed by a further decrease in 2025 (Google 26.6\%, DuckDuckGo 27.7\%). This downward trend is consistent across engines and aligns with the intuitive expectation that earlier cutoff dates allow a longer window for retrieved pages to accumulate post-cutoff updates, resulting in higher leakage rates.

\section{Conclusion}
% We demonstrate that date-filtered web search fails to prevent temporal leakage in LLM forecasting evaluation. Our analysis of 393 questions and 38,879 retrieved URLs reveals pervasive leakage that artificially inflates accuracy, improving the Brier score from a baseline of 0.24 to 0.10. Validated by our LLM-as-judge methodology, these findings indicate that current filtering is insufficient and call for stronger safeguards, such as evaluation on frozen, time-stamped web snapshots

We demonstrate that date-filtered web search fails to prevent temporal leakage in retrospective forecasting evaluation. Auditing Google Search's \texttt{before:} filter across 393 resolved forecasting questions and DuckDuckGo's date-range filter across 389, we find that major post-cutoff information appears in at least one retrieved page for 71\% of questions on Google and 81\% on DuckDuckGo, and the answer is directly revealed for 41\% and 55\%, respectively. In a downstream forecasting experiment on Google-audited documents, access to leaky content inflates apparent accuracy, reducing the Brier score from 0.24 with leak-free documents to 0.10 with strong-to-direct leakage. Leakage also varies with cutoff recency: URL-level post-cutoff rates are highest for 2021--2022 cutoffs (>46\%) and decline for more recent ones, consistent with older pages having more time to accumulate post-cutoff updates. We further characterize four recurring leakage mechanisms: page updates, related-content modules, absence-based signals, and unreliable metadata. These findings indicate that date-restricted search on these major engines is insufficient for credible retrospective evaluation and call for stronger safeguards such as evaluation on frozen, time-stamped web snapshots.

\section*{Limitations}
\label{sec:limitations}
Our study is subject to several limitations. First, our audit focuses on Google Search and DuckDuckGo as representative major search engines and on the Metaculus forecasting platform. While leakage appears across both engines, leakage prevalence and mechanisms may differ for other retrieval systems. Second, our leakage detection and forecasting experiments both use the same \texttt{gpt-oss-120b} model, raising the possibility of shared interpretation biases. The judge and forecaster may read a document the same way, including when that reading is wrong. While our human-LLM agreement and high per-class F1 for direct-leakage cases suggest the judge's labels are reliable, future audits of this kind would benefit from judges drawn from different models. Third, our document processing pipeline uses Maximal Marginal Relevance (MMR) to handle long texts, which risks omitting dispersed leakage signals in excluded chunks from our analysis. Finally, we diagnose and quantify the leakage problem but do not experimentally evaluate mitigation strategies such as Wayback Machine retrieval or frozen snapshot databases. Such comparisons are valuable future work.

\section*{Acknowledgments}
\label{sec:Acknowledgments}
We used LLMs only for sentence-level polishing (clarity, wording, and grammatical corrections) and limited implementation assistance for small refactors or boilerplate. All LLM-suggested changes were reviewed, edited as needed, and verified by the authors.

% \input{Sections/07-ethical_considerations}  optional

\bibliography{custom}
\appendix

\section{Methodology Detail}
\label{sec:appendix-mothodology}

\subsection{Document Processing}
\label{sec:document-processing}
For long documents, we apply Maximal Marginal Relevance (MMR) \citep{10.1145/290941.291025} to select the most relevant content while maintaining diversity. We chunk documents into 256-token segments and select up to 30 chunks using the Qwen-0.6B \citep{qwen3technicalreport} embedding model with $\lambda = 0.7$ (balancing relevance and diversity via cosine similarity). Documents under 7,680 tokens (256 $\times$ 30) threshold are passed in full. This is particularly important for the forecasting experiments, where models receive multiple documents (e.g., several score-3 documents and one score-4 document). See Appendix~\ref{sec:appendix-prompts} for complete prompts.

\subsection{LLM-Human Agreement}
\label{sec:appendix-llm-human-agreement}
\begin{figure}[h]
  \includegraphics[width=\columnwidth]{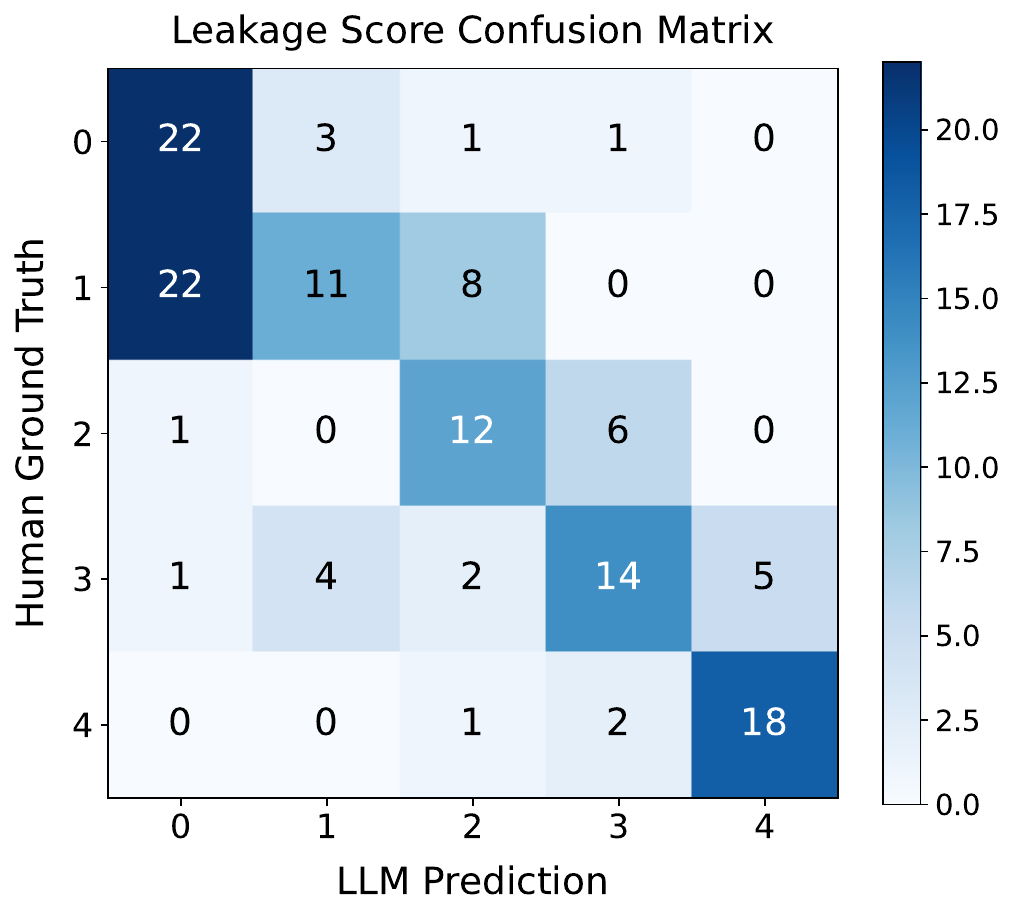}
  \caption{Confusion matrix of human-LLM score}
  \label{fig:human_llm_score_confusion_matrix}
\end{figure}

Two annotators scored 134 documents using the same rubric. Figure~\ref{fig:human_llm_score_confusion_matrix} shows the confusion matrix.

% \FloatBarrier
\subsection{Example Metaculus Question}
\label{sec:example-question}
Below is an example resolved forecasting question from our dataset, illustrating the structure of a Metaculus question.

\begin{description}[style=unboxed, leftmargin=0pt, itemsep=1pt, topsep=2pt]
\item[Title:] Will an additional state join NATO before 2024?
\item[ID:] 8549
\item[Background:] Since its founding, the admission of new member states has increased the alliance from the original 12 countries to 30. The most recent member state to be added to NATO was North Macedonia on 27 March 2020\ldots{} Members agreed that their aim is to reach or maintain the target defense spending of at least 2\% of their GDP by 2024.
\item[Open Time:] 2021-11-18T15:00:00Z
\item[Actual Close Time:] 2023-04-04T14:30:00Z
\item[Actual Resolve Time:] 2023-04-04T14:30:00Z
\item[Status:] resolved
\item[Type:] binary
\item[Resolution Criteria:] The question will resolve positively if, at any time between January 1, 2021 to January 1, 2024, any state formally joins NATO. This will be resolved based on an official statement by NATO, for example by the new state being included in the member list on NATO's official website. If a current NATO member fragments into two or more successor states and one or more of these join NATO, this will not count toward a positive resolution.
\item[Resolution:] yes
\item[Fine Print:] None.
\end{description}

\section{Case Study URLs}
\label{sec:appendix-case-studies}

\begin{description}[style=unboxed, leftmargin=0pt, itemsep=4pt]
\item[Direct Page Updates:] \url{https://missilethreat.csis.org/north-korea-missile-launches-1984-present/}
\item[Related Content Module:] \url{https://thecipherbrief.com/nuclear-deterrence-and-assurance-in-east-asia}
\item[Absence-Based Signal:] \url{https://cnn.com/interactive/2025/06/world/us-iran-conflict-timeline-dg/}
\item[Unreliable Metadata:] \url{https://cfr.org/election2020/candidate-tracker}
\end{description}
% \FloatBarrier

\section{Prompts}
\label{sec:appendix-prompts}
\subsection{Query Generation Prompt}
\label{sec:prompt-query}
\begin{lstlisting}
You are an expert in using search engines and writing search keywords.
We will breakdown and decompose the user query into {number_of_queries} distinct search queries.
Generate {number_of_queries} distinct search queries that would help gather comprehensive information about this topic.
Each query should focus on a different aspect or perspective.
The queries should be precise, concise, friendly for search engines (not complete sentences), SEO-aware, and relevant to the original query.
Generate queries in the user's native language, do not do any translation. 
Return only the queries as a JSON array.

Your response must be a valid JSON array of strings, wrapped with <JSON> and </JSON>.

Example output:
{json_queries_example}

Now, generate the queries based on the user query:
{user_query}
\end{lstlisting}

\subsection{LLM-as-a-Judge Data Leakage Prompt}
\label{app:prompt-judge}

\begin{lstlisting}
You are an expert Data Contamination Auditor.
Your job is to detect if the given text chunk leaks information that originates past the given Information Cutoff Date and how significant this leakage is in answering the given forecasting question.

Question: {question}
Background Info: {background}
Resolution Criteria: {resolution_criteria}
Resolved answer: {resolved_answer}
Information Cutoff Date: {information_cutoff_date}

Post-Cutoff information: Any event, data point, or entity that did not exist or was not public knowledge prior to the Information Cutoff Date.


Judge Process
Step 1:
Scan the text chunk for post-cutoff information, i.e. information that originates after the Information Cutoff Date ({information_cutoff_date}). These can include logical leaks of events/entities that didn't exist pre-cutoff or retroactive revelations where there is a post-cutoff disclosure of previously secret pre-cutoff events. 

Step 2:
Score the post-cutoff information given the Leakage Score Rubric below. 

Leakage Score Rubric:

0 - Noise (Irrelevant or no post-cutoff information)
Assign 0 if no post-cutoff info exists or post-cutoff info exists but is irrelevant to the question topic and would not change a forecaster's probability estimate at all.

1 - Topical
Assign 1 if the post-cutoff info is related to the question topic but would not help or change a forecaster's answer and reasoning in determining the question's answer.

2 - Weak Signal
Assign 2 if the post-cutoff info is relevant to the question and provides a vague/minor directional signal towards the question's resolution. These are often vague trend statements with little or no quantitative detail.

3 - Major or Partial Direct Signal
Assign 3 if the post-cutoff info provides a major component/signal toward the question's resolution. This could involve having direct/decisive data for a partial component of the question or data from a later date than the desired question date that can be used to strongly infer the resolved answer.


4 - Full Decisive/Direct Leakage
Assign 4 if the post-cutoff info directly reveals the resolved answer whether by explicitly stating it or providing data allowing for a certain logical/mathematical calculation of the answer.

Sometimes what isn't mentioned (absence of information) is important for forecasting. Use absence of information only in clearly comprehensive or high-quality reports that are relevant to the question. Otherwise, score based on what is stated, not omissions. When absence of information is reasonable, it is scored a 3 at most, as we do not want to assume a certain 4. 


Example to help anchor scoring:
Ex Question: Will X and Y countries ratify treaty Z by 2024?
Example Cutoff Date: January 15, 2021
'Copyright 2024' is assigned a 0. Irrelevant post-cutoff metadata. 
'In 2024, the Treaty Z Secretariat renamed its Working Group 2 to the 'Implementation Committee.'' is assigned a 1. Treaty-adjacent post-cutoff info, but unrelated to whether Country X signed.
'(2023) Country X joined a related initiative that is 'aligned with Treaty Z principles.' is assigned a 2.
'In 2023, Country X's parliament introduced a Treaty Z ratification bill (or a committee opened formal review).' is assigned a 3. Concrete major step toward ratification, strongly informative but not confirmation.
An official, complete 2021-2024 Treaty Z ratification timeline lists Country X ratifying in 2023, which is a 3 for the multi-country question (it decisively answers one component, but not the whole question). The same comprehensive timeline does not list Country Y, which is also a 3 absence-based signal for non-ratification.
The Treaty Z Depositary's official registry (updated 2024) lists Country X ratifying in 2023 and explicitly states Country Y is not eligible to ratify Treaty Z (so no ratification by Y is possible). This is assigned a 4.

Output Format:

Provide your response as a valid JSON object wrapped with <JSON> and </JSON> containing exactly these keys:


<JSON>
{{
  "reasoning": "First, provide all the post-cutoff info found. Then assign a score for each piece of post-cutoff info found based on the Leakage Score Rubric and explain your rationale.",  "contains_post_cutoff_info": boolean,
  "leakage_score": integer (0-4) #Apply the highest leakage score given for the post-cutoff info found here
}}
</JSON>

Text chunk to evaluate:
{context}\end{lstlisting}
\subsection{Binary Forecasting Prompt}
\label{app:prompt-forecast}
\begin{lstlisting}
You are a professional forecaster interviewing for a job.
The question's forecasting window begins on {open_time}.
Your interview question is:
{title}

Question background:
{background}

This question's outcome will be determined by the specific criteria below. These criteria have not yet been satisfied:
{resolution_criteria}

{fine_print}

Summary Research Report:
Warning: These snippets are from an automated search engine. They may contain irrelevant info, conflicting data, or headlines that do not tell the full story. They may also have ambigious dates. You must evaluate them critically and check specific numbers against the resolution criteria.
{summary_report}

Before answering you write:
(a) The time left from now until the resolution date. Consider the forecasting window of when it began and the resolution date.
(b) The status quo outcome if nothing changed.
(c) A brief description of a scenario that results in a No outcome.
(d) A brief description of a scenario that results in a Yes outcome.

You write your rationale remembering that good forecasters put extra weight on the status quo outcome since the world changes slowly most of the time.

The last thing you write is your final answer. You must write the probability of the "Yes" outcome only. Format it exactly as: "Probability: ZZ%", 0-100
\end{lstlisting}

% \FloatBarrier

\end{document}